\documentclass[11pt]{article}

\usepackage[preprint]{acl}

\usepackage{arydshln}
\usepackage{amssymb}
\usepackage{times}
\usepackage{latexsym}
\usepackage{pifont}
\usepackage{booktabs}
\usepackage{multirow}
\usepackage[table]{xcolor}
\usepackage{makecell}
\usepackage{mwe}
\usepackage[export]{adjustbox} 

\usepackage[T1]{fontenc}

\usepackage[utf8]{inputenc}

\usepackage{microtype}

\usepackage{inconsolata}

\usepackage{graphicx}

\newcommand{\cmark}{\ding{51}}%
\newcommand{\xmark}{\ding{55}}%

\newcommand{\ours}{Z3D}

\title{Z3D: Zero-Shot 3D Visual Grounding from Images}

\author{\bfseries Nikita Drozdov\textsuperscript{1} \quad Andrey Lemeshko\textsuperscript{2} \quad Nikita Gavrilov\textsuperscript{1} \\ \bfseries Anton Konushin\textsuperscript{1} \quad Danila Rukhovich\textsuperscript{3} \quad Maksim Kolodiazhnyi\textsuperscript{1\dag} \\
\textsuperscript{1}Lomonosov Moscow State University\quad
\textsuperscript{2}Higher School of Economics \\
\textsuperscript{3}M3L Lab, Institute of Mechanics, Armenia}

\begin{document}
\maketitle

\begin{figure*}
\centering
\includegraphics[width=0.985\linewidth]{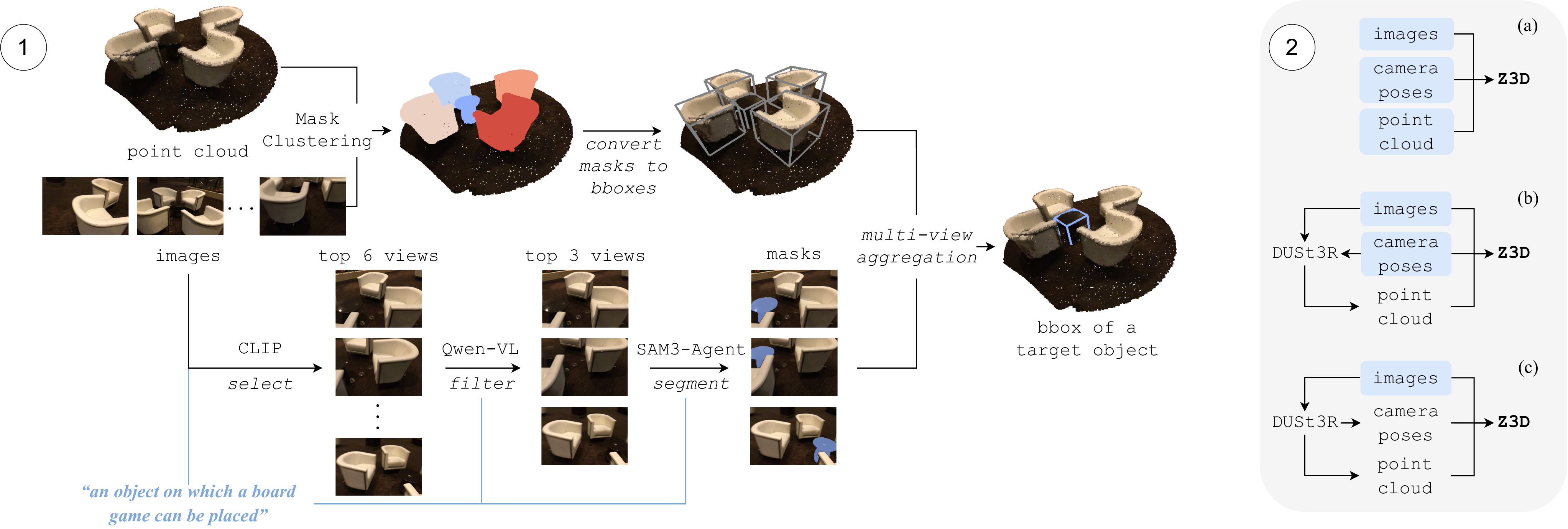}
\caption{(1) Inference pipeline of \ours~ given point cloud inputs. \ours~ leverages MaskClustering~\cite{yan2024maskclustering} to predict class-agnostic 3D bounding boxes from a point cloud and images. The most relevant frames are chosen based on their CLIP embeddings and further filtered with VLM. Object masks are annotated using SAM3-Agent~\cite{carion2025sam3} and lifted to 3D. The final bounding box is selected from MaskClustering proposals through a mask voting scheme. (2) Our approach flexibly accommodates different input modalities, including images, camera poses, and depth maps. When camera poses and depth maps are unavailable, they are estimated with DUSt3R~\cite{wang2024dust3r}.}
\label{fig:scheme}
\end{figure*}

\begin{abstract}
\let\thefootnote\relax\footnotetext{\textsuperscript{\dag}Corresponding author: kolodyazhniyma@my.msu.ru}\footnotetext{Code available at \url{https://github.com/col14m/z3d}}
3D visual grounding (3DVG) aims to localize objects in a 3D scene based on natural language queries. In this work, we explore zero-shot 3DVG from multi-view images alone, without requiring any geometric supervision or object priors. We introduce \ours, a universal grounding pipeline that flexibly operates on multi-view images while optionally incorporating camera poses and depth maps. We identify key bottlenecks in prior zero-shot methods causing significant performance degradation and address them with (i) a state-of-the-art zero-shot 3D instance segmentation method to generate high-quality 3D bounding box proposals and (ii) advanced reasoning via prompt-based segmentation, which utilizes full capabilities of modern VLMs. Extensive experiments on the ScanRefer and Nr3D benchmarks demonstrate that our approach achieves state-of-the-art performance among zero-shot methods.
\end{abstract}

\section{Introduction}

3D visual grounding (3DVG) seeks to localize target objects in a scene based on natural language descriptions. It is a fundamental capability for embodied AI, robotics, and human–scene interaction, where agents must reason jointly over language, visual appearance, and spatial structure. 

\paragraph{3DVG from point clouds}
is the most natural and well-studied scenario. The recent emergence of LLM allows reducing the 3D labeling burden, leveraging generalization capabilities of large models to eliminate the need for 3D supervision. Earlier methods, such as Vil3DRel~\cite{chen2022vil3drel} and MiKASA~\cite{chang2024mikasa}, and their recent follow-ups, SceneVerse~\cite{jia2024sceneverse}, LIBA~\cite{wang2025liba}, ROSS3D~\cite{wang2025ross3d}, LlaVA-3D~\cite{zhu2025llava-3d}, Video-3D LLM~\cite{zheng2025video-3d-llm}, GPT4Scene~\cite{qi2025gpt4scene}, and MPEC~\cite{wang2025mpec}, rely on full supervision, with both 3D bounding boxes and texts being exposed to the model during the training phase. ZSVG3D~\cite{yuan2024zsvg3d}, CSVG~\cite{yuan2024csvg}, SeeGround~\cite{li2025seeground}, LaSP~\cite{mi2025lasp}, EaSe~\cite{mi2025ease}, SPAZER~\cite{jin2025spazer} do not require texts for training but still exploit annotated 3D bounding boxes. The training-free, zero-shot approach is represented with LLM-Grounder~\cite{yang2024llm-grounder} and VLM-Grounder~\cite{xu2024vlm-grounder}. Both of them rely on proprietary VLMs but still use non-generative language models, such as BERT and CLIP, in critical parts of the pipeline, which severely limits their performance. Overcoming this weakness, we achieve up to +40\% in accuracy over prior state-of-the-art. Besides, we investigate another cause of poor performance of prior zero-shot approaches: the insufficient quality of candidate object proposals. Using state-of-the-art zero-shot 3D instance segmentation method, we generate high-quality proposals that serve as a strong basis for further VLM reasoning.

\paragraph{3DVG from images}
Most existing zero-shot methods assume access to explicit 3D representations, such as point clouds, depth maps, or pre-built scene reconstructions, which restricts their applicability in real-world settings. Fully supervised SPAR~\cite{zhang2025spar} and VG LLM~\cite{zheng2025vg-llm} claim to be image-based, but both require ground truth camera pose: SPAR uses them during the test phase, while VG LLM is exposed to camera poses during the training. Zero-shot 3DVG is addressed with modern VLMs, e.g., Qwen3-VL~\cite{qwen3-vl} and Seed1.5-VL~\cite{guo2025seed1.5-vl}, but only from single-view images, which is a critical limitation, since we aim for scene-level understanding.  In this work, we study zero-shot 3D visual grounding from multi-view images alone and present \ours{}, a universal grounding pipeline that operates on images and can optionally incorporate camera poses and depth maps when available. 

Our contribution is twofold:
\begin{itemize}
\item we improve components of the existing VLM-based 3DVG pipeline, achieving state-of-the-art results in 3DVG from point clouds;
\item we extend our method to handle various inputs, thus being the first to address 3DVG in a zero-shot setting from images only.
\end{itemize}

\section{Proposed Method}

\subsection{3DVG With Depth}

3DVG implies estimating a 3D bounding box of a target object in a scene given a query in natural language. Existing zero-shot methods rely on VLMs to handle fuzzy indirect references and require images to proceed. The target object might be visible in a subset of frames, so the task of selecting the most relevant views arises naturally. Then, the target object must be located in those views and lifted to 3D space. Respectively, the VLM-based 3DVG workflow can be broadly decomposed into (i) view selection, (ii) 2D object segmentation, and (iii) 2D-to-3D lifting. The only existing zero-shot baseline showing reasonable performance, VLM-Grounder~\cite{xu2024vlm-grounder}, follows this paradigm; in \ours, we propose modifications of each step that push the quality to the state-of-the-art level.

\paragraph{View selection.}

Processing all images of a scene with VLM is time-consuming, so optimizations are inevitable. VLM-Grounder packs images in a grid to minimize the number of calls and iteratively narrows the search scope until finding the best views. In contrast, we employ a two-stage strategy to efficiently identify informative observations. First, views are preselected using CLIP so that the six most similar frames for a given query pass the filter. Then, VLM selects the best three views. Therefore, the search space is reduced using a lightweight model and simple selection strategy, while the accuracy benefits of a more powerful but computationally expensive approach are retained.

\paragraph{2D object segmentation.}

VLM-Grounder uses a combination of Grounding DINO~\cite{ren2024grounding-dino-1.5} and SAM to localize and segment objects in a frame, respectively. This standard pipeline, powered by BERT~\cite{devlin2019bert}, is better tailored for open-vocabulary segmentation with explicit object descriptions, while less specific 3DVG prompts might cause performance degradation. To overcome this limitation, \ours~ employs a SAM3-Agent~\cite{carion2025sam3} for zero-shot high-quality object segmentation. Guided by VLM reasoning, the agent iteratively generates and refines segmentation prompts, enabling precise instance extraction without geometric supervision or object priors.

\paragraph{2D-to-3D lifting.}

Each object mask can be lifted into 3D space, giving a partial point cloud of an object. VLM-Grounder simply unites all partial point clouds and encloses them with a 3D bounding box to create a proposal; hence, outliers have a large impact on the estimated size and shape of an object, making the whole procedure prone to noise. Differently, we first obtain class-agnostic 3D object proposals and use 2D object masks to select the best one. To this end, we leverage MaskClustering~\cite{yan2024maskclustering}, a zero-shot 3D instance segmentation method that takes point clouds as inputs and produces object 3D masks, which we convert into 3D bounding boxes. Segmentation masks from the top-3 views are lifted to 3D and matched against the MaskClustering proposals. A proposal with the highest 3D IoU with a mask gets one vote; the final proposal is the most voted candidate, or, in ambiguous cases, the one voted by a mask in the frame with the highest CLIP and VLM relevance scores.

\begin{table*}
    \centering
    \resizebox{\linewidth}{!}{
    \begin{tabular}{lllcccccccc}
        \toprule 
        & \multirow{2}{*}{Method} & \multirow{2}{*}{Venue} & \multicolumn{2}{c}{Supervision} & \multicolumn{2}{c}{Unique} & \multicolumn{2}{c}{Multiple} & \multicolumn{2}{c}{Overall} 
        \\
        & & & bboxes & texts & Acc@0.25 & Acc@0.5 & Acc@0.25 & Acc@0.5 & Acc@0.25 & Acc@0.5 \\
        \midrule
        \multicolumn{3}{l}{\textit{Images + camera poses + depths}} \\
        & LLaVA-3D & ICCV'25 & \cmark & \cmark & - & - & - & -  & 50.1 & 42.7 \\
        & Video-3D LLM & CVPR'25 & \cmark & \cmark & 86.6 & 77.0 & 50.9 & 45.0 & 57.9 & 51.2 \\
        & ROSS3D & ICCV'25 & \cmark & \cmark & 87.2 & 77.4 & 54.8 & 48.9   & 61.1 & 54.4 \\
        & LIBA & AAAI'25 & \cmark & \cmark & 88.8 & 74.3 & 54.4 & 44.4 & 59.6 & 49.0 \\
        & GPT4Scene & - & \cmark & \cmark & 90.3 & 83.7 & 56.4 & 50.9 & 62.6 & 57.0 \\  
        \cdashline{2-11}
        & ZSVG3D & CVPR'24 & Mask3D & \xmark & 63.8 & 58.4 & 27.7 & 24.6 & 36.4 & 32.7 \\
        & CSVG & BMVC'25 & Mask3D & \xmark & 68.8 & 61.2 & 38.4 & 27.3 & 49.6 & 39.8 \\
        & SeeGround & CVPR'25 & Mask3D & \xmark & 75.7 & 68.9 & 34.0 & 30.0 & 44.1 & 39.4 \\
        \rowcolor{blue!10} & \ours~ & - & Mask3D & \xmark & \textbf{82.3} & \textbf{74.8} & \textbf{51.5} & \textbf{45.7} & \textbf{58.9} & \textbf{52.7} \\
        \cdashline{2-11}
        & OpenScene & CVPR'23 & \xmark & \xmark &  20.1 & 13.1 & 11.1 & 4.4 & 13.2 & 6.5 \\
        & LLM-Grounder & ICRA'24 & \xmark & \xmark & - & -  & - & - & 17.1 & 5.3 \\
        \rowcolor{blue!10} & \ours & - & \xmark & \xmark & \textbf{73.9} & \textbf{64.0} & \textbf{47.8} & \textbf{40.3} & \textbf{54.2} & \textbf{46.0} \\
        \midrule
        \multicolumn{3}{l}{\textit{Images + camera poses}} \\
        & SPAR & NIPS'25 & \cmark & \cmark & - & - & - & - & 31.9 & 12.4 \\
        \cdashline{2-11}
        \rowcolor{blue!4} & DUSt3R $\rightarrow$ SeeGround & - & \xmark & \xmark & 44.5 & 27.4 & 21.1 & 12.6 & 26.8 & 16.2 \\
        \rowcolor{blue!10} & DUSt3R $\rightarrow$ \ours & - & \xmark & \xmark & \textbf{56.7} & \textbf{32.0} & \textbf{38.4} & \textbf{22.6} & \textbf{42.8} & \textbf{24.8} \\
        \midrule
        \multicolumn{3}{l}{\textit{Images}} \\
        & VG LLM & NIPS'25 & \cmark & \cmark & - & - & - & - & 41.6 & 14.9 \\
        \cdashline{2-11}
        \rowcolor{blue!4} & DUSt3R $\rightarrow$ SeeGround & - & \xmark & \xmark & 35.2 & 17.5 & 13.9 & 5.2 & 19.0 & 8.2 \\
        \rowcolor{blue!10} & DUSt3R $\rightarrow$ \ours & - & \xmark & \xmark & \textbf{42.7} & \textbf{21.9} & \textbf{27.5} & \textbf{10.1} & \textbf{31.2} & \textbf{12.9} \\
        \bottomrule
    \end{tabular}
    }
    \caption{Results on ScanRefer. Queries are categorized as \textit{``Unique''}, with only one object of the target class in the scene, or \textit{``Multiple''}, with other objects of the same class. Results of our method are blue, our baselines are marked with a lighter hue.}
    \label{tab:results-scanrefer}
\end{table*}

\begin{table}
    \centering
    \resizebox{0.975\linewidth}{!}{
    \begin{tabular}{rlccccc}
        \toprule
        & Method & Easy & Hard & Dep. & Indep. & Overall \\
        \midrule
        \multicolumn{7}{l}{\textit{Fully supervised}} \\
        & MiKASA & 69.7 & 59.4 & 65.4 & 64.0 & 64.4 \\
        & ViL3DRel & 70.2 & 57.4 & 62.0 & 64.5 & 64.4 \\
        & SceneVerse & 72.5 & 57.8 & 56.9 & 67.9 & 64.9 \\
        & MPEC & - & - & - & - & 66.7 \\
        \midrule
        \multicolumn{7}{l}{\textit{Zero-shot (use gt object class)}} \\
        & CSVG & 67.1 & 51.3 & 53.0 & 62.5 & 59.2 \\
        & EaSe & - & - & - & - & 67.8 \\
        & Transcrib3D & 79.7 & 60.3 & 60.1 & 75.4 & 70.2 \\
        \midrule
        \multicolumn{7}{l}{\textit{Zero-shot}} \\
        & ZSVG3D & 46.5 & 31.7 & 36.8 & 40.0 & 39.0 \\
        & SeeGround & 54.5 & 38.3 & 42.3 & 48.2 & 46.1 \\
        & LaSP  & 60.7 & 45.3 & 49.2  & 54.7 & 52.9 \\
        & EaSe  & - & - &  - & - & 52.9 \\
        & SPAZER & 62.4 & 46.9 & 49.9 & 56.8 & 54.3 \\ 
        \rowcolor{blue!10} & \ours & \textbf{62.6} & \textbf{47.5} & \textbf{50.7} & \textbf{57.1} & \textbf{54.8} \\
        \bottomrule
    \end{tabular}
    }
    \caption{Results of methods using depths on Nr3D, specified by the query type: \textit{``Easy''} (one distractor) \textit{``Hard''} (multiple distractors), \textit{``View-Dependent''} or \textit{``View-Independent''} based on viewpoint requirements for grounding.}
    \label{tab:results-nr3d}
\end{table}

\begin{table}
\centering
\resizebox{0.975\linewidth}{!}{
\begin{tabular}{lccc}
\toprule
Module & Acc@$0.25$ & Acc@$0.5$  \\
\midrule
MaskClustering~\cite{yan2024maskclustering} & 32.0 & 27.6 \\
 + CLIP + SAM3-Agent & 51.0  & 42.8 \\
 + VLM view selection & 53.0  & 44.8 \\ 
\rowcolor{blue!10} + multi-view aggregation & \textbf{54.2} & \textbf{46.0} \\
\bottomrule
\end{tabular}
}
\caption{Ablation study of model components on ScanRefer, with images, camera poses, and depths as inputs.}
\label{tab:ablation-model-components}
\end{table}

\subsection{3DVG From Images}

When depths or point clouds are unavailable, we bridge the gap between sole visual inputs and real geometry with a 3D reconstruction method.
Specifically, we use DUSt3R~\cite{wang2024dust3r}: with its ability to seamlessly handle omnimodal inputs, it fits perfectly into both images-only and images + camera poses scenarios. With DUSt3R, our processing pipeline remains purely zero-shot, since, contrary to some latest methods~\cite{wang2025vggt} it was not trained on ScanNet~\cite{dai2017scannet}. Given images, DUSt3R returns dense depth maps and infers poses when they are not available. The depths are then fused into a TSDF volume using ground truth or predicted camera poses. Finally, a point cloud is extracted using the marching cubes algorithm.

\section{Experiments}

We evaluate our approach on the ScanRefer~\cite{chen2020scanrefer} and Nr3D~\cite{achlioptas2020referit3d} benchmarks. ScanRefer annotates ScanNet scenes with over 51K human-written query–object pairs, where the goal is to localize the target object by predicting its 3D bounding box from scene point clouds and language queries. Following standard practice, we report Acc@0.25 and Acc@0.5, defined as the percentage of predictions whose 3D IoU with ground truth exceeds 0.25 and 0.5, respectively.
The Nr3D dataset contains 41K language queries over ScanNet scenes and provides ground-truth 3D bounding boxes without class labels. The task is to select the most relevant candidate object, which is evaluated using top-1 accuracy.

\paragraph{3DVG with depths}

3DVG methods that use depth, point clouds, or other sources of spatial information represent the most extensively studied setting. These approaches can be categorized based on their exposure to 3D bounding boxes providing geometric supervision: (i) methods trained with ground-truth bounding boxes (e.g., approaches using Mask3D proposals), (ii) methods provided with bounding boxes at inference time (as in the Nr3D benchmark), and (iii) methods that are not exposed to bounding boxes at any stage. When using Mask3D as a proposal generator, \ours~ demonstrates substantial improvements over prior methods that are exposed to bounding boxes in the training set (Tab.~\ref{tab:results-scanrefer}, row 9). In the inference-time 3D bounding box setting, \ours~ achieves state-of-the-art top-1 accuracy on Nr3D (Tab.~\ref{tab:results-nr3d}), indicating that the gain stems not only from proposal quality but also from the effectiveness of the remaining components of our pipeline. Finally, in the purely zero-shot setting without any bounding-box supervision, \ours~ significantly outperforms all competitors, achieving an absolute improvement of +38.7 Acc@0.5 over OpenScene on ScanRefer (Tab.~\ref{tab:results-scanrefer}, row 12).

\paragraph{3DVG from images}

For image-based 3DVG, both with and without camera poses, existing approaches are fully supervised; therefore, we report their results for reference only. To establish a meaningful baseline, we combine DUSt3R with a state-of-the-art point cloud–based 3DVG method. Specifically, we adopt SeeGround~\cite{li2025seeground}, which is highly competitive in depth-aware settings. While the original SeeGround uses Mask3D to generate proposals, in this series of experiments, we replace it with MaskClustering to keep the whole pipeline zero-shot. As shown in Tab.~\ref{tab:results-scanrefer} (rows 15, 18), \ours~ consistently outperforms SeeGround on DUSt3R reconstructions, demonstrating that its advantages are preserved regardless of the reconstruction approach. Notably, \ours~ establishes a new state of the art in the posed-images setting, surpassing the previous state-of-the-art fully supervised method, SPAR~\cite{zhang2025spar}.

\paragraph{Ablation study}

To quantify the contribution of each component, we conduct an ablation study by progressively building our pipeline from a simple baseline. The original MaskClustering method, designed for open-vocabulary 3D instance segmentation, already achieves 27.6 Acc@0.5. Incorporating CLIP-based view selection (top-1 view) and SAM3-Agent for object segmentation increases performance to 42.8. When the number of views selected by CLIP is increased to 6, and followed by top-1 view selection with VLM, accuracy further improves to 44.8. Finally, aggregating predictions across the top-3 views selected by the VLM results in the best performance, reaching 46.0.

\section{Conclusion}

We presented \ours, a universal pipeline for zero-shot 3D visual grounding, with a particular focus on the grounding from multi-view images alone. By identifying proposal quality and underutilization of VLMs as key bottlenecks in prior methods, we addressed these limitations through the integration of zero-shot 3D instance segmentation and VLM reasoning. Our approach flexibly accommodates different input modalities, including multi-view images, camera poses, and depth maps. Evaluations on ScanRefer and Nr3D demonstrate that \ours~ achieves state-of-the-art performance among zero-shot approaches across multiple settings. We hope this work encourages further research on image-based and supervision-free 3D visual grounding, paving the way toward more practical and scalable 3D scene understanding systems.

\section*{Limitations}

While introducing advanced VLM reasoning about the selected frames, our method still uses CLIP to pre-select frame candidates and therefore can be limited by CLIP's ability to analyze complex concepts from subtle cues rather than direct descriptions. Moreover, in image-only scenarios, the performance of our method heavily depends on the quality of underlying 3D reconstruction. While DUSt3R is known to perform robustly on ScanNet captures, the similar quality is not guaranteed for other scenes. In terms of performance, one of the processing bottlenecks of \ours~ is MaskClustering, which adds a significant computation overhead; the component-wise time analysis can be found in the supplementary materials.

\bibliography{custom}

\begin{thebibliography}{31}
\providecommand{\natexlab}[1]{#1}

\bibitem[{Achlioptas et~al.(2020)Achlioptas, Abdelreheem, Xia, Elhoseiny, and Guibas}]{achlioptas2020referit3d}
Panos Achlioptas, Ahmed Abdelreheem, Fei Xia, Mohamed Elhoseiny, and Leonidas Guibas. 2020.
\newblock Referit3d: Neural listeners for fine-grained 3d object identification in real-world scenes.
\newblock In \emph{European conference on computer vision}, pages 422--440. Springer.

\bibitem[{Bai et~al.(2025)Bai, Cai, Chen et~al.}]{qwen3-vl}
Shuai Bai, Yuxuan Cai, Ruizhe Chen, and 1 others. 2025.
\newblock Qwen3-vl technical report.
\newblock \emph{arXiv preprint arXiv:2511.21631}.

\bibitem[{Carion et~al.(2025)Carion, Gustafson, Hu, Debnath, Hu, Suris, Ryali, Alwala, Khedr, Huang et~al.}]{carion2025sam3}
Nicolas Carion, Laura Gustafson, Yuan-Ting Hu, Shoubhik Debnath, Ronghang Hu, Didac Suris, Chaitanya Ryali, Kalyan~Vasudev Alwala, Haitham Khedr, Andrew Huang, and 1 others. 2025.
\newblock Sam 3: Segment anything with concepts.
\newblock \emph{arXiv preprint arXiv:2511.16719}.

\bibitem[{Chang et~al.(2024)Chang, Wang, Pagani, and Stricker}]{chang2024mikasa}
Chun-Peng Chang, Shaoxiang Wang, Alain Pagani, and Didier Stricker. 2024.
\newblock Mikasa: Multi-key-anchor \& scene-aware transformer for 3d visual grounding.
\newblock In \emph{Proceedings of the IEEE/CVF Conference on Computer Vision and Pattern Recognition}, pages 14131--14140.

\bibitem[{Chen et~al.(2020)Chen, Chang, and Nie{\ss}ner}]{chen2020scanrefer}
Dave~Zhenyu Chen, Angel~X Chang, and Matthias Nie{\ss}ner. 2020.
\newblock Scanrefer: 3d object localization in rgb-d scans using natural language.
\newblock In \emph{European conference on computer vision}, pages 202--221. Springer.

\bibitem[{Chen et~al.(2022)Chen, Guhur, Tapaswi, Schmid, and Laptev}]{chen2022vil3drel}
Shizhe Chen, Pierre-Louis Guhur, Makarand Tapaswi, Cordelia Schmid, and Ivan Laptev. 2022.
\newblock Language conditioned spatial relation reasoning for 3d object grounding.
\newblock \emph{Advances in neural information processing systems}, 35:20522--20535.

\bibitem[{Dai et~al.(2017)Dai, Chang, Savva, Halber, Funkhouser, and Nie{\ss}ner}]{dai2017scannet}
Angela Dai, Angel~X Chang, Manolis Savva, Maciej Halber, Thomas Funkhouser, and Matthias Nie{\ss}ner. 2017.
\newblock Scannet: Richly-annotated 3d reconstructions of indoor scenes.
\newblock In \emph{Proceedings of the IEEE conference on computer vision and pattern recognition}, pages 5828--5839.

\bibitem[{Devlin et~al.(2019)Devlin, Chang, Lee, and Toutanova}]{devlin2019bert}
Jacob Devlin, Ming-Wei Chang, Kenton Lee, and Kristina Toutanova. 2019.
\newblock Bert: Pre-training of deep bidirectional transformers for language understanding.
\newblock In \emph{Proceedings of the 2019 conference of the North American chapter of the association for computational linguistics: human language technologies, volume 1 (long and short papers)}, pages 4171--4186.

\bibitem[{Guo et~al.(2025)Guo, Wu, Zhu, Leng, Shi, Chen, Fan, Wang, Jiang, Wang et~al.}]{guo2025seed1.5-vl}
Dong Guo, Faming Wu, Feida Zhu, Fuxing Leng, Guang Shi, Haobin Chen, Haoqi Fan, Jian Wang, Jianyu Jiang, Jiawei Wang, and 1 others. 2025.
\newblock Seed1. 5-vl technical report.
\newblock \emph{arXiv preprint arXiv:2505.07062}.

\bibitem[{Jia et~al.(2024)Jia, Chen, Yu, Wang, Niu, Liu, Li, and Huang}]{jia2024sceneverse}
Baoxiong Jia, Yixin Chen, Huangyue Yu, Yan Wang, Xuesong Niu, Tengyu Liu, Qing Li, and Siyuan Huang. 2024.
\newblock Sceneverse: Scaling 3d vision-language learning for grounded scene understanding.
\newblock In \emph{European Conference on Computer Vision}, pages 289--310. Springer.

\bibitem[{Jin et~al.(2025)Jin, Tu, Liao, Sun, Luo, Liu, and Tao}]{jin2025spazer}
Zhao Jin, Rong-Cheng Tu, Jingyi Liao, Wenhao Sun, Xiao Luo, Shunyu Liu, and Dacheng Tao. 2025.
\newblock Spazer: Spatial-semantic progressive reasoning agent for zero-shot 3d visual grounding.
\newblock \emph{arXiv preprint arXiv:2506.21924}.

\bibitem[{Li et~al.(2025)Li, Li, Kong, Yang, and Liang}]{li2025seeground}
Rong Li, Shijie Li, Lingdong Kong, Xulei Yang, and Junwei Liang. 2025.
\newblock Seeground: See and ground for zero-shot open-vocabulary 3d visual grounding.
\newblock In \emph{Proceedings of the Computer Vision and Pattern Recognition Conference}, pages 3707--3717.

\bibitem[{Mi et~al.(2025{\natexlab{a}})Mi, Wang, Wang, Chen, and Pang}]{mi2025ease}
Boyu Mi, Hanqing Wang, Tai Wang, Yilun Chen, and Jiangmiao Pang. 2025{\natexlab{a}}.
\newblock Evolving symbolic 3d visual grounder with weakly supervised reflection.
\newblock \emph{arXiv preprint arXiv:2502.01401}.

\bibitem[{Mi et~al.(2025{\natexlab{b}})Mi, Wang, Wang, Chen, and Pang}]{mi2025lasp}
Boyu Mi, Hanqing Wang, Tai Wang, Yilun Chen, and Jiangmiao Pang. 2025{\natexlab{b}}.
\newblock Language-to-space programming for training-free 3d visual grounding.
\newblock In \emph{Proceedings of the 2025 Conference on Empirical Methods in Natural Language Processing}, pages 3844--3864.

\bibitem[{Qi et~al.(2025)Qi, Zhang, Fang, Wang, and Zhao}]{qi2025gpt4scene}
Zhangyang Qi, Zhixiong Zhang, Ye~Fang, Jiaqi Wang, and Hengshuang Zhao. 2025.
\newblock Gpt4scene: Understand 3d scenes from videos with vision-language models.
\newblock \emph{arXiv preprint arXiv:2501.01428}.

\bibitem[{Ren et~al.(2024)Ren, Jiang, Liu, Zeng, Liu, Gao, Huang, Ma, Jiang, Chen et~al.}]{ren2024grounding-dino-1.5}
Tianhe Ren, Qing Jiang, Shilong Liu, Zhaoyang Zeng, Wenlong Liu, Han Gao, Hongjie Huang, Zhengyu Ma, Xiaoke Jiang, Yihao Chen, and 1 others. 2024.
\newblock Grounding dino 1.5: Advance the" edge" of open-set object detection.
\newblock \emph{arXiv preprint arXiv:2405.10300}.

\bibitem[{Teed and Deng(2021)}]{teed2021droid-slam}
Zachary Teed and Jia Deng. 2021.
\newblock Droid-slam: Deep visual slam for monocular, stereo, and rgb-d cameras.
\newblock \emph{Advances in neural information processing systems}, 34:16558--16569.

\bibitem[{Wang et~al.(2025{\natexlab{a}})Wang, Zhao, Wang, Fan, Zhang, and Zhang}]{wang2025ross3d}
Haochen Wang, Yucheng Zhao, Tiancai Wang, Haoqiang Fan, Xiangyu Zhang, and Zhaoxiang Zhang. 2025{\natexlab{a}}.
\newblock Ross3d: Reconstructive visual instruction tuning with 3d-awareness.
\newblock \emph{arXiv preprint arXiv:2504.01901}.

\bibitem[{Wang et~al.(2025{\natexlab{b}})Wang, Chen, Karaev, Vedaldi, Rupprecht, and Novotny}]{wang2025vggt}
Jianyuan Wang, Minghao Chen, Nikita Karaev, Andrea Vedaldi, Christian Rupprecht, and David Novotny. 2025{\natexlab{b}}.
\newblock Vggt: Visual geometry grounded transformer.
\newblock In \emph{Proceedings of the Computer Vision and Pattern Recognition Conference}, pages 5294--5306.

\bibitem[{Wang et~al.(2024)Wang, Leroy, Cabon, Chidlovskii, and Revaud}]{wang2024dust3r}
Shuzhe Wang, Vincent Leroy, Yohann Cabon, Boris Chidlovskii, and Jerome Revaud. 2024.
\newblock Dust3r: Geometric 3d vision made easy.
\newblock In \emph{Proceedings of the IEEE/CVF Conference on Computer Vision and Pattern Recognition}, pages 20697--20709.

\bibitem[{Wang et~al.(2025{\natexlab{c}})Wang, Jia, Zhu, and Huang}]{wang2025mpec}
Yan Wang, Baoxiong Jia, Ziyu Zhu, and Siyuan Huang. 2025{\natexlab{c}}.
\newblock Masked point-entity contrast for open-vocabulary 3d scene understanding.
\newblock In \emph{Proceedings of the Computer Vision and Pattern Recognition Conference}, pages 14125--14136.

\bibitem[{Wang et~al.(2025{\natexlab{d}})Wang, Li, ZY, and Wang}]{wang2025liba}
Yuan Wang, Ya-Li Li, WU~Eastman ZY, and Shengjin Wang. 2025{\natexlab{d}}.
\newblock Liba: Language instructed multi-granularity bridge assistant for 3d visual grounding.
\newblock In \emph{Proceedings of the AAAI Conference on Artificial Intelligence}, volume~39, pages 8114--8122.

\bibitem[{Xu et~al.(2024)Xu, Huang, Wang, Chen, Pang, and Lin}]{xu2024vlm-grounder}
Runsen Xu, Zhiwei Huang, Tai Wang, Yilun Chen, Jiangmiao Pang, and Dahua Lin. 2024.
\newblock Vlm-grounder: A vlm agent for zero-shot 3d visual grounding.
\newblock \emph{arXiv preprint arXiv:2410.13860}.

\bibitem[{Yan et~al.(2024)Yan, Zhang, Zhu, and Wang}]{yan2024maskclustering}
Mi~Yan, Jiazhao Zhang, Yan Zhu, and He~Wang. 2024.
\newblock Maskclustering: View consensus based mask graph clustering for open-vocabulary 3d instance segmentation.
\newblock In \emph{Proceedings of the IEEE/CVF Conference on Computer Vision and Pattern Recognition}, pages 28274--28284.

\bibitem[{Yang et~al.(2024)Yang, Chen, Qian, Madaan, Iyengar, Fouhey, and Chai}]{yang2024llm-grounder}
Jianing Yang, Xuweiyi Chen, Shengyi Qian, Nikhil Madaan, Madhavan Iyengar, David~F Fouhey, and Joyce Chai. 2024.
\newblock Llm-grounder: Open-vocabulary 3d visual grounding with large language model as an agent.
\newblock In \emph{2024 IEEE International Conference on Robotics and Automation (ICRA)}, pages 7694--7701. IEEE.

\bibitem[{Yuan et~al.(2024{\natexlab{a}})Yuan, Zhang, Li, and Stiefelhagen}]{yuan2024csvg}
Qihao Yuan, Jiaming Zhang, Kailai Li, and Rainer Stiefelhagen. 2024{\natexlab{a}}.
\newblock Solving zero-shot 3d visual grounding as constraint satisfaction problems.
\newblock \emph{arXiv preprint arXiv:2411.14594}.

\bibitem[{Yuan et~al.(2024{\natexlab{b}})Yuan, Ren, Feng, Zhao, Cui, and Li}]{yuan2024zsvg3d}
Zhihao Yuan, Jinke Ren, Chun-Mei Feng, Hengshuang Zhao, Shuguang Cui, and Zhen Li. 2024{\natexlab{b}}.
\newblock Visual programming for zero-shot open-vocabulary 3d visual grounding.
\newblock In \emph{Proceedings of the IEEE/CVF Conference on Computer Vision and Pattern Recognition}, pages 20623--20633.

\bibitem[{Zhang et~al.(2025)Zhang, Chen, Zhou, Xu, Huang, Mei, Chen, Yuan, Cai, Huang et~al.}]{zhang2025spar}
Jiahui Zhang, Yurui Chen, Yanpeng Zhou, Yueming Xu, Ze~Huang, Jilin Mei, Junhui Chen, Yu-Jie Yuan, Xinyue Cai, Guowei Huang, and 1 others. 2025.
\newblock From flatland to space: Teaching vision-language models to perceive and reason in 3d.
\newblock \emph{arXiv preprint arXiv:2503.22976}.

\bibitem[{Zheng et~al.(2025{\natexlab{a}})Zheng, Huang, Li, and Wang}]{zheng2025vg-llm}
Duo Zheng, Shijia Huang, Yanyang Li, and Liwei Wang. 2025{\natexlab{a}}.
\newblock Learning from videos for 3d world: Enhancing mllms with 3d vision geometry priors.
\newblock \emph{arXiv preprint arXiv:2505.24625}.

\bibitem[{Zheng et~al.(2025{\natexlab{b}})Zheng, Huang, and Wang}]{zheng2025video-3d-llm}
Duo Zheng, Shijia Huang, and Liwei Wang. 2025{\natexlab{b}}.
\newblock Video-3d llm: Learning position-aware video representation for 3d scene understanding.
\newblock In \emph{Proceedings of the Computer Vision and Pattern Recognition Conference}, pages 8995--9006.

\bibitem[{Zhu et~al.(2025)Zhu, Wang, Zhang, Pang, and Liu}]{zhu2025llava-3d}
Chenming Zhu, Tai Wang, Wenwei Zhang, Jiangmiao Pang, and Xihui Liu. 2025.
\newblock Llava-3d: A simple yet effective pathway to empowering lmms with 3d capabilities.
\newblock In \emph{Proceedings of the IEEE/CVF International Conference on Computer Vision}, pages 4295--4305.

\end{thebibliography}

\appendix


\begin{table*}[t!]
\centering
\resizebox{\linewidth}{!}{
\begin{tabular}{llcccccccc}
    \toprule 
    \multirow{2}{*}{Method} & \multirow{2}{*}{Venue} & \multicolumn{2}{c}{Supervision} & \multicolumn{2}{c}{Unique} & \multicolumn{2}{c}{Multiple} & \multicolumn{2}{c}{Overall} \\
    & & bboxes & texts & Acc@0.25 & Acc@0.5 & Acc@0.25 & Acc@0.5 & Acc@0.25 & Acc@0.5 \\
    \midrule
    ZSVG3D & CVPR'24 & Mask3D & \xmark & 55.3 & 55.3 & 25.6 & 25.6 & 31.2 & 31.2 \\
    SeqVLM & ACMMM'25  & Mask3D & \xmark & 77.3 & 72.7  & 47.8 & 41.3 & 55.6 & 49.6 \\
    SPAZER  & NIPS'25 & Mask3D & \xmark & 80.9 & 72.3 & 51.7 & 43.4 & 57.2 & 48.8 \\
    \rowcolor{blue!10} \ours~ & - & Mask3D & \xmark  & \textbf{87.9} & \textbf{81.8} & \textbf{51.7} & \textbf{44.6} & \textbf{61.2} & \textbf{54.4} \\
    \midrule
    LLM-Grounder & ICRA'24 & \xmark & \xmark & 12.1	& 4.0	& 11.7	& 5.2 & 12.0	& 4.4  \\
    VLM-Grounder  & CoRL'24 & \xmark & \xmark  & 66.0 & 29.8  & 48.3 & 33.5 & 51.6 & 32.8\\
    \rowcolor{blue!10} \ours & - & \xmark & \xmark & \textbf{78.8} & \textbf{71.2}  & \textbf{50.5} & \textbf{44.6} & \textbf{58.0} & \textbf{51.6}\\
    \bottomrule
\end{tabular}
}
\caption{Evaluations of 3DVG from point clouds on 250 scenes subset from ScanRefer.}
\label{tab:scanrefer250}
\end{table*}

\begin{table*}
\centering
\resizebox{0.7\linewidth}{!}{
\begin{tabular}{llccccc}
\toprule
Method & VLM & Easy & Hard & Dep. & Indep. & Overall \\ 
\midrule
SeeGround & Qwen2-VL-72B & 51.5 & 37.7 & 44.8 & 45.5 & 45.2 \\
VLM-Grounder & GPT-4o & 55.2 & 39.5 & 45.8 & 49.4 & 48.0 \\
SeqVLM & Doubao-1.5-vision-pro & 58.1 & 47.4 & 51.0 & 54.5 & 53.2 \\
SPAZER & Qwen2.5-VL-72B & 60.3 & 50.9 & 54.2 & 57.1 & 56.0 \\ 
\rowcolor{blue!10} \ours & Qwen2.5-VL-72B & 66.9 & 44.7 & 58.3 & 55.8 & \textbf{56.8} \\ 
\midrule
\rowcolor{blue!10} \ours & Qwen3-VL-8B-Thinking & 59.6 & 47.4 & 54.2 & 53.9 & 54.0 \\ 
\rowcolor{blue!10} \ours & Qwen3-VL-30B-Thinking & 66.2 & 47.4 & 57.3 & 57.8 & 57.6 \\ 
\rowcolor{blue!10} \ours & Qwen3-VL-235B-Thinking & 68.4 & 47.4 & 58.3 & 59.1 & \textbf{58.8} \\ 
\bottomrule
\end{tabular}
}
\caption{Evaluations of 3DVG from point clouds on 250 scenes subset from Nr3D.}
\label{tab:nr3d250}
\end{table*}

\section{Quantitative Results}

Some recent methods follow the alternative evaluation protocol, proposed in VLM-Grounder~\cite{xu2024vlm-grounder}. This protocol implies testing on 250-scene subsets of ScanRefer and Nr3D rather than their full validation splits. The results on ScanRefer and Nr3D are reported in Tab.~\ref{tab:scanrefer250} and \ref{tab:nr3d250}, respectively; clearly, \ours~ scores the best in both benchmarks. \ours~ shines in the pure zero-shot scenario (w/o access to ground truth 3D bounding boxes), achieving +18.8 Acc@0.5 w.r.t. VLM-Grounder on ScanRefer. On Nr3D, our method outperforms previous state-of-the-art SPAZER~\cite{jin2025spazer} using the same Qwen2.5-VL-72B, and even beats VLM-Grounder based on much more powerful proprietary GPT-4o.

\section{Ablation Studies}

In this Section, all results are reported on 250-scenes subsets.

\paragraph{VLM size} In Tab.~\ref{tab:nr3d250}, we vary the size of Qwen3-VL-Thinking serving as our VLM reasoner, and report the quality achieved with each model size. Even with a 30B model, \ours~ outperforms prior methods, and using a larger 235B model pushes the quality even further.

\paragraph{Mask3D vs. MaskClustering} The key difference between MaskClustering and Mask3D is that the first is a pure training-free approach, while the second is trained with ground truth 3D bounding box annotations. In Tab.~\ref{tab:scanrefer250}, we demonstrate that even with less exposure to the training data, \ours~ outperforms methods that source object proposals with Mask3D. When using Mask3D, \ours~ shows +2.8 Acc@0.5 on ScanRefer w.r.t. the best competing approach in the respective category.

\paragraph{Number of images} In image-base scenarios, the number of input images is a crucial aspect of the model's performance. According to the experiments on ScanRefer, the more images, the better (Tab.~\ref{tab:ablation-n-frames}). Since the reconstruction quality is highly correlated with the coverage of a scene, this conclusion can be expected. Existing approaches use a comparable number of images, e.g., VLM-Grounder takes up to 60 images and VG LLM uses 24 images. Still, after the view selection procedure, all methods reason based on fewer images: 3 in \ours, 7 in VLM-Grounder, or 6 in VG LLM. 

\paragraph{DUSt3R vs. DROID-SLAM} To investigate how dependent is our pipeline on the reconstruction quality, we replace DUSt3R with DROID-SLAM~\cite{teed2021droid-slam}. According to Tab.~\ref{tab:ablation-slam}, this leads to a massive drop of scores: apparently, DROID-SLAM cannot deliver the sufficient quality produce to localize and recognize 3D objects reliably.

\paragraph{Inference time} We measure inference time component-wise and report the performance in Tab.~\ref{tab:time}. The most time-consuming part of our pipeline is MaskClustering, while other components are executed relatively fast. Overall, \ours~ is on par with the zero-shot baseline VLM-Grounder.

\begin{table}
\centering
\resizebox{0.975\linewidth}{!}{%
\begin{tabular}{llllcc}
\toprule
\textbf{Method} & \multicolumn{3}{c}{\textbf{Step}} & \textbf{Time (s)} & \textbf{Total (s)} \\ \midrule
\multirow{5}{*}{\ours} & \multicolumn{3}{c}{MaskClustering} & 56.3 & \multirow{5}{*}{61.0} \\
& \multicolumn{3}{c}{CLIP view selection} & 0.001 & \\
& \multicolumn{3}{c}{VLM view selection} & 1.5 & \\
& \multicolumn{3}{c}{SAM3-Agent} & 2.8 & \\
& \multicolumn{3}{c}{multi-view aggregation}  & 0.4 & \\
\midrule
\multirow{3}{*}{SPAZER} & \multicolumn{3}{c}{view selection} & 5.2 & \multirow{3}{*}{23.5} \\
& \multicolumn{3}{c}{candidate object screening}  & 8.5 & \\
& \multicolumn{3}{c}{3D-2D decision-making} & 9.8 & \\ 
\midrule
VLM-Grounder & \multicolumn{3}{c}{-} & - & 50.3 \\ 
\bottomrule
\end{tabular}
}
\caption{Inference time of each step in \ours.}
\label{tab:time}
\end{table}

\begin{table}
\centering
\resizebox{0.975\linewidth}{!}{
\begin{tabular}{cccccc}
\toprule
\multirow{2}{*}{\# Images} & \multicolumn{2}{c}{Images} & \multicolumn{2}{c}{Images + camera poses}  \\
& Acc@0.25 & Acc@0.5 & Acc@0.25 & Acc@0.5 \\
\midrule
15 & 20.3 & 8.5 & 32.6 & 17.1 \\
45 & \textbf{30.0} & \textbf{12.7} & \textbf{41.1} & \textbf{24.0} \\
\bottomrule
\end{tabular}
}
\caption{Results of \ours{} from images with and without poses on ScanRefer 250-scenes subset with varying number of images.}
\label{tab:ablation-n-frames}
\end{table}

\begin{table}
\centering
\resizebox{0.85\linewidth}{!}{
\begin{tabular}{lcc}
\toprule
Method & Acc@0.25 & Acc@0.5 \\
\midrule
DROID-SLAM $\rightarrow$ \ours & 14.3 & 4.8 \\
DUSt3R $\rightarrow$ \ours & \textbf{30.0} & \textbf{12.7} \\
\bottomrule
\end{tabular}
}
\caption{Results of \ours{} from images on ScanRefer 250-scenes subset with different pose estimation methods.}
\label{tab:ablation-slam}
\end{table}

\section{Qualitative Results}

\begin{figure*}
\centering \small
\resizebox{\linewidth}{!}{
\begin{tabular}{ccccc}
\multirow{2}{*}{Ground Truth} & \multicolumn{3}{c}{\ours{} predictions from} & \multirow{2}{*}{Text prompt} \\
& Images + Poses + Depths & Images + Poses & Images \\
\includegraphics[width=0.20\linewidth, valign=c]
{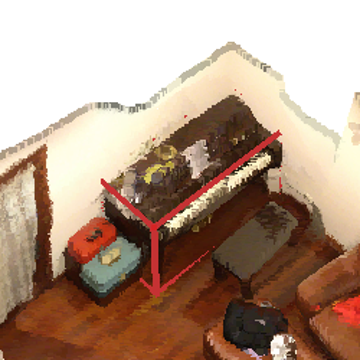} & 
\includegraphics[width=0.20\linewidth, valign=c]{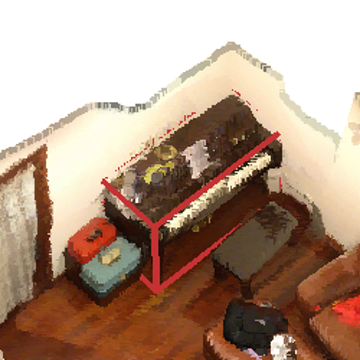} & 
\includegraphics[width=0.20\linewidth, valign=c]
{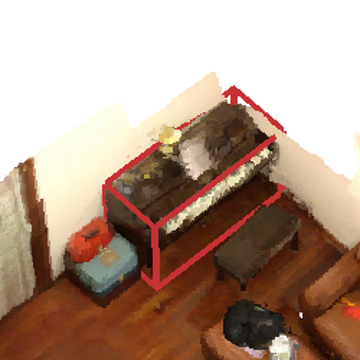} & \includegraphics[width=0.20\linewidth, valign=c]
{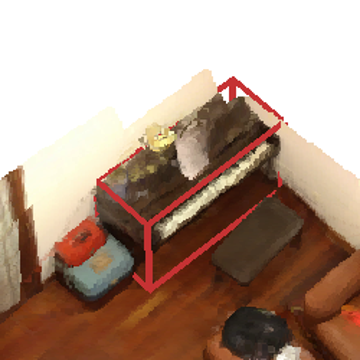} & \makecell[l]{\itshape The brown piano is to the \\ \itshape right of the double doors. \\ \itshape There are a red and blue \\ \itshape case to the left of the piano.} \\
\includegraphics[width=0.20\linewidth, valign=c]{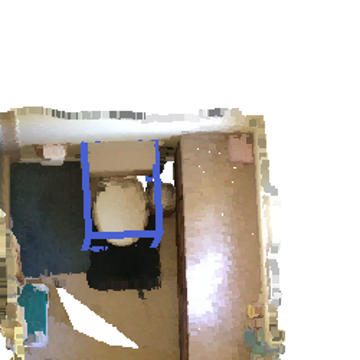} & 
\includegraphics[width=0.20\linewidth, valign=c]{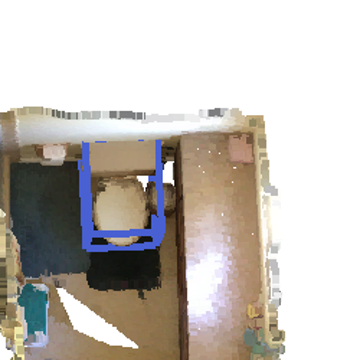} & 
\includegraphics[width=0.20\linewidth, valign=c]{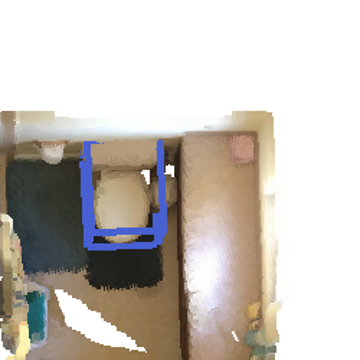} & 
\includegraphics[width=0.20\linewidth, valign=c]{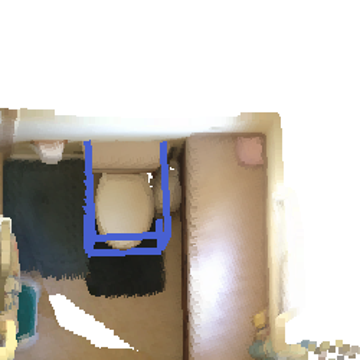} & \makecell[l]{\itshape The toilet is in the back of \\ \itshape the room. it is to the right \\ \itshape of the toilet paper and \\ \itshape to the left of the sink.} \\
\includegraphics[width=0.20\linewidth, valign=c]{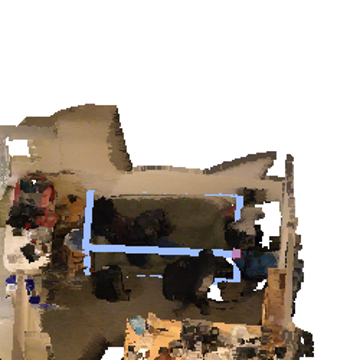} & 
\includegraphics[width=0.20\linewidth, valign=c]{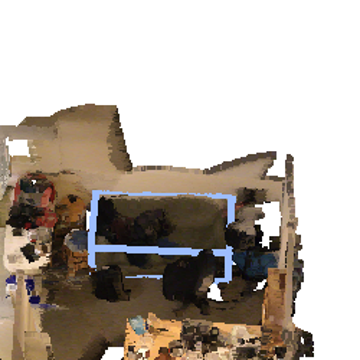} & 
\includegraphics[width=0.20\linewidth, valign=c]{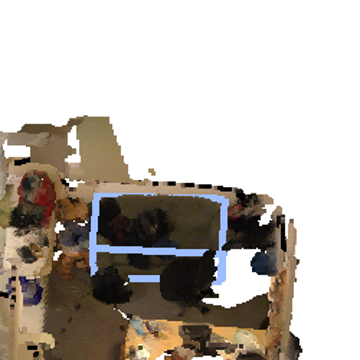} & 
\includegraphics[width=0.20\linewidth, valign=c]{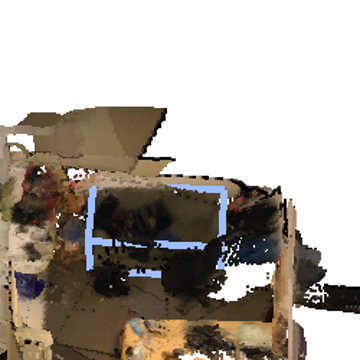} & \makecell[l]{\itshape The couch has two stools \\ \itshape to its left and a black chair \\ \itshape in front. The couch is dark \\ \itshape green and has two seats.} \\
\end{tabular}
}
\caption{Qualitative results of \ours{} on ScanRefer.}
\label{fig:scanrefer}
\end{figure*}

\begin{figure}
\centering \scriptsize
\setlength{\tabcolsep}{1pt}
\resizebox{\linewidth}{!}{
\begin{tabular}{ccc}
Ground Truth & \ours & Text prompt \\
\includegraphics[width=0.20\linewidth, valign=c]
{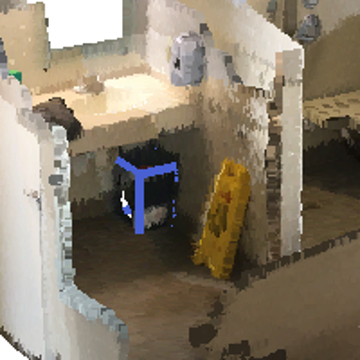} & 
\includegraphics[width=0.20\linewidth, valign=c]{images/nr3d/scene0084_00_gt.png} & 
\makecell[l]{\itshape The trash can below the \\ \itshape hand sanitizer and next \\ \itshape to the wet floor sign.} \\ \addlinespace[2pt]
\includegraphics[width=0.20\linewidth, valign=c]{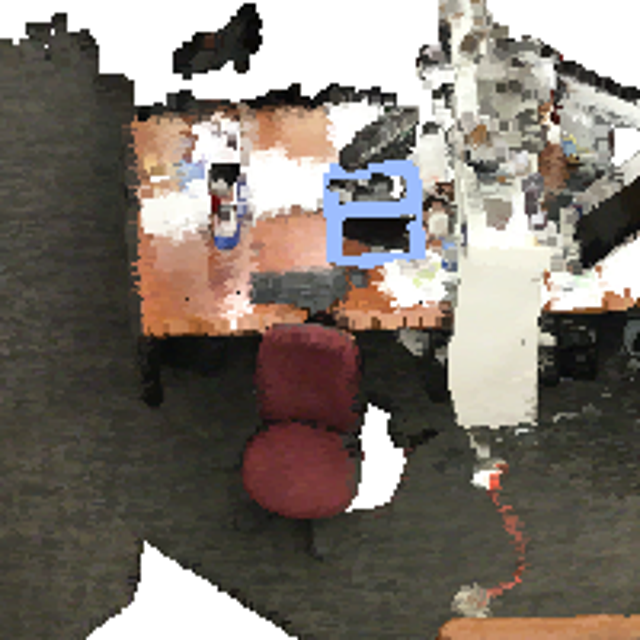} & 
\includegraphics[width=0.20\linewidth, valign=c]{images/nr3d/scene0329_00_gt.png} & 
\makecell[l]{\itshape The monitor at the \\ \itshape desk with the red chair \\ \itshape facing the wrong way} \\ \addlinespace[2pt]
\includegraphics[width=0.20\linewidth, valign=c]{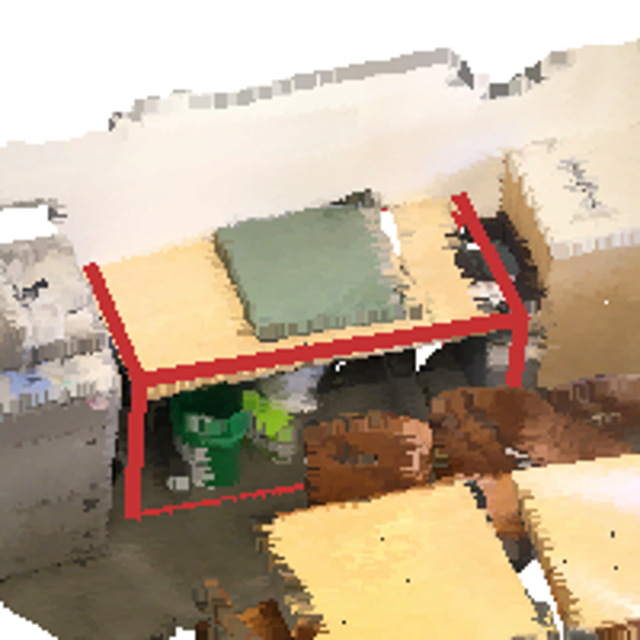} & 
\includegraphics[width=0.20\linewidth, valign=c]{images/nr3d/scene0552_00_gt.png} & 
\makecell[l]{\itshape The table that is next \\ \itshape to the wall and has a green \\ \itshape bucket underneath it.}
\end{tabular}
}
\caption{Qualitative results of \ours~ on Nr3D.}
\label{fig:nr3d}
\end{figure}

\begin{figure}
\centering \scriptsize
\setlength{\tabcolsep}{1pt}
\resizebox{\linewidth}{!}{
\begin{tabular}{ccc}
Ground Truth & \ours & Text prompt \\
\includegraphics[width=0.20\linewidth, valign=c]{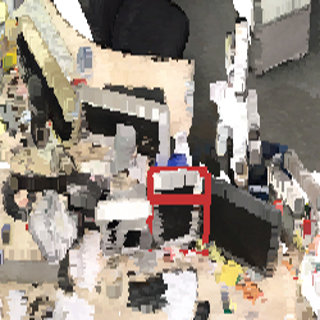} & 
\includegraphics[width=0.20\linewidth, valign=c]{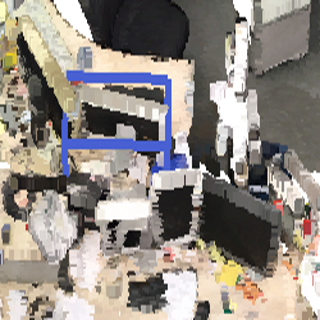} & 
\makecell[l]{\itshape In a row of three monitors, \\ \itshape the middle monitor.} \\ \addlinespace[2pt]
\includegraphics[width=0.20\linewidth, valign=c]{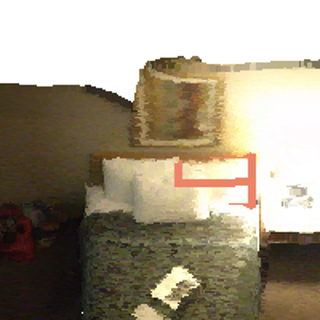} & 
\includegraphics[width=0.20\linewidth, valign=c]{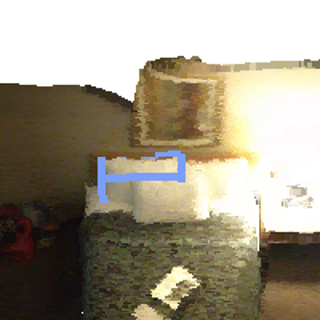} & 
\makecell[l]{\itshape The bed next to the dresser, \\ \itshape it is the pillow in the back, \\ \itshape closest to the nightstand.}
\end{tabular}
}
\caption{Failure cases of \ours~ on Nr3D.}
\label{fig:failure}
\end{figure}

\paragraph{ScanRefer} Fig.~\ref{fig:scanrefer} depicts \ours~ predictions on ScanRefer from all types of inputs: images solely, images with poses, and images with poses and depths. Comparison on the same scene shows how additional inputs contribute to the quality.

\paragraph{Nr3D} Predictions on Nr3D given images with poses and depths are shown in Fig.~\ref{fig:nr3d}. Nr3D benchmark provides ground truth 3D bounding boxes, from which the only one should be selected as an answer; accordingly, predicted boxes is strictly equal to ground truth ones if the guess is correct. 

\paragraph{Failure cases} We analyzed failure cases and identified a typical pattern. As can be observed in Fig.~\ref{fig:failure}, our model sometimes fails to select the object in the presence of multiple similar objects in a scene: monitors (top row), pillows (bottom row).

\end{document}